# Full waveform inversion method based on diffusion model

Liu Caiyun[1] Pei Siyang[1] Yu Qingfeng[2] Xiong Jie[2]

1) School of Information and Mathematics, Yangtze University, Jingzhou, Hubei Province 434023, China

2) School of Electronic Information and Electrical Engineering, Yangtze University, Jingzhou, Hubei Province 434023, China

**Abstract:** Seismic full-waveform inversion is a core technology for obtaining high-resolution subsurface model parameters. However, its highly nonlinear characteristics and strong dependence on the initial model often lead to the inversion process getting trapped in local minima. In recent years, generative diffusion models have provided a way to regularize full-waveform inversion by learning implicit prior distributions. However, existing methods mostly use unconditional diffusion processes, ignoring the inherent physical coupling relationship between velocity and density and other physical properties. This paper proposes a full-waveform inversion method based on conditional diffusion model regularization. By improving the backbone network structure of the diffusion model, two-dimensional density information is introduced as a conditional input into the U-Net network. Experimental results show that the full-waveform inversion method based on the conditional diffusion model significantly improves the resolution and structural fidelity of the inversion results, and exhibits stronger stability and robustness when dealing with complex situations. This method effectively utilizes density information to constrain the inversion and has good practical application value.

Before the 1970s and 1980s, earthquake inversion mainly relied on reflection coefficients and travel time information, and tomographic inversion was carried out based on linear or weakly nonlinear approximations[1], which could only recover the macroscopic velocity structure. Norton[2] verified the feasibility of iterative algorithms such as the steepest descent method and the conjugate gradient method in nonlinear reflection data inversion, laying the foundation for subsequent research. In the 1980s, Lailly[3] derived the gradient formula of model parameters based on the wave equation; Tarantola[4] proposed time-domain full waveform inversion based on generalized least squares, getting rid of the limitation of only using reflection wave information, and laying the theoretical foundation for time-domain full waveform inversion. In the 1990s, Pratt et al.[5] proposed the frequency domain FWI, which improved the engineering practicality of the method. In 2009, Shin and Lee et al.[6] proposed the Laplace domain full waveform inversion, which reduced the dependence on the initial model by recovering the macroscopic velocity structure. In 2012, Xu et al.[7] proposed the reflection wave waveform inversion method, which achieved good results in the application of actual seismic data on land.

To alleviate the problems of strong nonlinearity, heavy dependence on the initial model and sensitivity to noise in the full waveform inversion process, researchers have gradually introduced regularization strategies to improve the stability and reliability of the inversion. In 1992, to solve the problem of model transition smoothing, Leonid Rudin et al.[8] proposed the TV regularization method, which maintains the model boundary structure by constraining the model gradient sparsity. On this basis, researchers have gradually introduced regularization strategies into full waveform inversion. In 2012, Felix J. Herrmann et al.[9] introduced compressed sensing theory into full waveform inversion, and reduced the computational cost by sparse representation and random sampling strategies. In 2017, Zhu et al.[10] proposed a sparse constraint full waveform inversion method based on dictionary learning to improve the model representation ability and enhance the inversion stability. In 2018, Tariq Alkhalifah et al.[11] proposed a full model wavenumber inversion method, which improves the inversion resolution by recovering the wavenumber information. In 2019, Huang et al.[12] proposed a robust full waveform inversion method based on dictionary learning to enhance the inversion's adaptability to noise. In 2021, Yue Li et al.[13] proposed a deep learning-assisted regularized full waveform inversion method, which improves the stability and accuracy of inversion by introducing well data velocity prior.

In recent years, the rise of artificial intelligence and deep learning technologies has provided new ideas for full waveform inversion. In 2020, Sheng Li et al.[14] proposed a seismic inversion based on deep neural networks (DNN) to learn the mapping between the data and model domains directly in a supervised manner, which improved the inversion efficiency. In 2020, Mosser et al.[15] proposed to introduce generative adversarial networks as geological priors into the stochastic full waveform inversion framework. By learning the probability distribution of the subsurface medium model, the inversion space is constrained, which improves the stability and geological rationality of the inversion results. In 2021, Zhang and Gao[16] proposed a deep learning full waveform inversion method based on seismic migration imaging results. The migration imaging results are used as network input, which alleviates the sensitivity of full waveform inversion to the initial model. In 2022, Tariq Alkhalifah et al.[17] proposed the MLReal method, which improves the generalization ability of machine learning models in actual seismic data by reducing the distribution difference between training data and real seismic data. In 2023, Yang and Ma[18] proposed the Generative Adversarial Network for Physical Information (FWIGAN), which combines physical constraints with adversarial learning to improve the accuracy and generalization ability of deep learning inversion results. With the rapid development of diffusion models in the field of image generation, Wang et al.[19] proposed a prior regularized full waveform inversion method based on diffusion models. By learning the probability distribution of the velocity model, they embedded it into the full waveform inversion optimization process, realizing the joint constraint of data fitting and generation prior, which effectively improved the stability of inversion.

Existing methods typically employ unconditional diffusion models to model velocity distributions, neglecting the intrinsic relationships between multiple physical properties. In real-world strata, velocity and density exhibit significant physical coupling, jointly influencing wavefield propagation characteristics. Therefore, this paper, building upon existing research, improves the diffusion network architecture by constructing a conditional diffusion model based on two-dimensional density constraints.

# 1 Method Principle

## 1.1 Full waveform inversion

Full Waveform Inversion (FWI) is a high-resolution seismic imaging and parameter inversion method based on the wave equation. The basic idea of FWI is to utilize the physical coupling relationship between subsurface medium parameters and the seismic wavefield, and reconstruct subsurface physical parameters by minimizing the difference between observed seismic data and numerical forward modeling data. Its workflow is shown in Figure 1. Compared with traditional travel-time inversion or pre-stack inversion methods, FWI can fully utilize the amplitude, phase, and multiple information in seismic records, thus offering higher resolution and stronger parameter recovery capabilities.

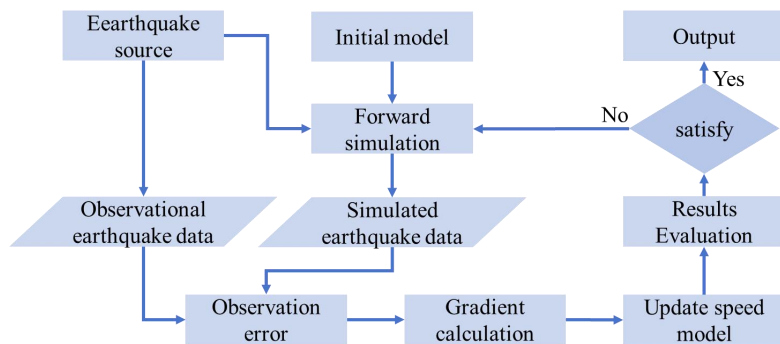

Fig. 1 Flow chart of full waveform inversion

Within the conventional framework, the objective function of FWI is typically defined as the L2 error between observed and simulated data:

$$J(m) = \frac{1}{2}\|d_{\text{obs}} - F(m)\|_2^2$$

Where $m$ represents the model parameters to be inverted, $d_{obs}$ is the observed seismic data, and $F(m)$ is the simulated data obtained from the forward modeling of the current model. This objective function focuses on the data fitting term and iteratively updates the model parameters by minimizing the data residuals.

Because seismic inversion suffers from ill-posedness, under limited observation conditions and noise interference, relying solely on data fitting terms for optimization often fails to yield a stable and unique solution, and may even lead to local minima. To enhance inversion stability and alleviate ill-posedness, a regularization term is typically introduced into the objective function to impose prior constraints on the model. The objective function after adding regularization can be expressed as::

$$J(m) = \frac{1}{2}\|d_{\text{obs}} - F(m)\|_2^2 + \lambda R(m)$$

Where $R(m)$ is the regularization term and $\lambda$ is the regularization weight coefficient, used to balance the contributions between the data fitting term and the prior constraint term.

## 1.2 Denoising Diffusion Probabilistic Model

The Denoising Diffusion Probabilistic Model (DDPM), or simply the diffusion model, is a type of probabilistic generation model that has been developed in recent years[20]. This model achieves the mapping between complex data distribution and simple Gaussian distribution by constructing a forward diffusion process that gradually adds noise and a corresponding reverse denoising generation process, thereby completing the modeling of high-dimensional data distribution. The basic process of the diffusion model is shown in Figure 2. The diffusion model includes two processes: the forward diffusion process and the reverse denoising process.

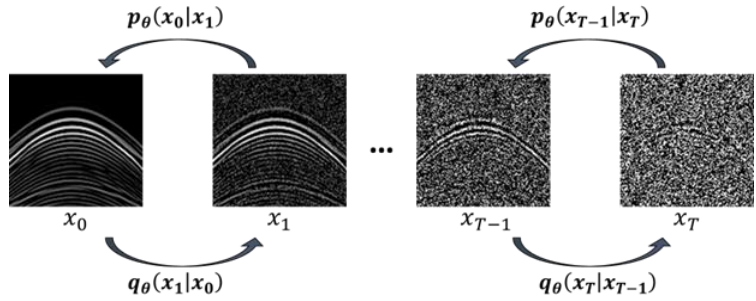

Fig. 2 Diffusion model process

In the forward diffusion process, Gaussian noise is gradually added to the original data $x_0$ to transform it into a noise variable $x_T$. This process can be represented as a Markov chain:

$$q(x_{1:T}|x_0) = \prod_{t=1}^{T} q(x_t|x_{t-1})$$

Each state transition in the process follows a Gaussian distribution:

$$q(x_t|x_{t-1}) \sim N(x_t; \sqrt{\alpha_t}x_{t-1}, (1-\alpha_t)I)$$

By continuously adding noise, the data distribution eventually approximates a standard Gaussian distribution. This allows us to obtain an explicit expression for any time step:

$$x_t = \sqrt{\bar{\alpha}_t}x_0 + \sqrt{1-\bar{\alpha}_t}\epsilon, \epsilon \sim N(0, I)$$

Among them, $\bar{\alpha}_t = \prod_{i=1}^{t} \alpha_t$。

In the reverse denoising stage, the model starts from Gaussian noise and gradually recovers the data distribution. This process learns the conditional probability distribution through a parameterized neural network:

$$p_\theta(x_{t-1}|x_t)$$

In actual training, to simplify the training process, the network is usually optimized by predicting noise. The neural network does not directly predict $x_{t-1}$ or the posterior mean, but instead predicts the noise component $\epsilon_\theta(x_t, t)$ contained in the sample at the current time step. The simplified training objective can be expressed as minimizing the mean square error between the predicted noise and the actual noise:

$$L_{simple} = E_{x_0,\epsilon}[\|\epsilon - \epsilon_\theta(x_t, t)\|^2]$$

## 1.3 Diffusion Regularized Full Waveform Inversion

Unlike traditional explicit regularization terms that rely on predefined mathematical forms, diffusion models learn the statistical regularities of a large number of velocity model samples to obtain implicit probabilistic expressions of subsurface media structures. They are no longer limited to smooth or sparse assumptions and can more naturally preserve non-smooth interfaces, multi-scale structures, and high wavenumber information in velocity models, thereby improving the resolution and structural fidelity of the inversion results.

Figure 3 shows the full waveform inversion process based on the diffusion model. The upper part of the figure shows the full waveform process based on the diffusion model. The black arrows represent the original unconditional inverse diffusion process, the red arrows represent the iterative process of traditional FWI, and the blue arrows represent the inverse process of the diffusion model. The lower part of the figure shows the specific update process at a single diffusion time step. Starting from a certain diffusion time step t and the initial velocity model, within each diffusion time step, several iterations of traditional FWI are first performed, and then the updated velocity model is corrected using the diffusion model. The worse the initial model used, the more diffusion steps are required.

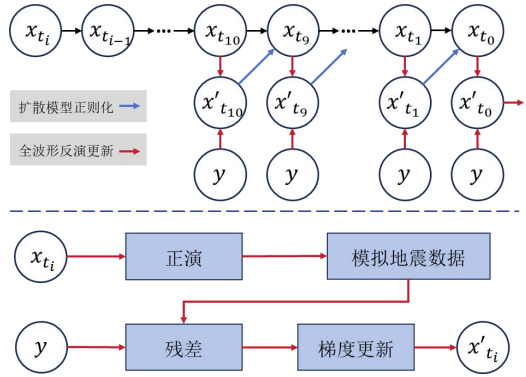

Fig. 3 Overall flow chart

The improved network structure is shown in Figure 4.

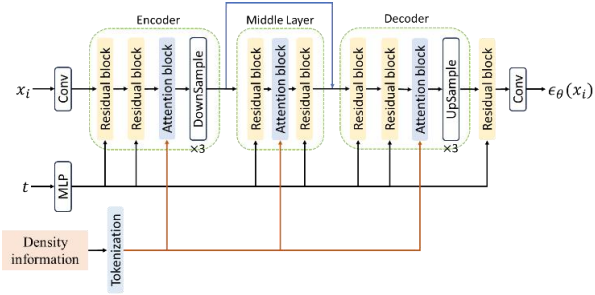

Fig. 4 Denoising network structure diagram

# 2  2. Model Experiment and Result Analysis

## 2.1 Pre-training of the diffusion model

This paper selects some underground velocity models from the OpenFWI dataset[21] as training data for the diffusion model to learn the statistical distribution characteristics of the underground medium velocity model and provide prior constraints for the subsequent full waveform inversion process. The OpenFWI dataset contains eight velocity models, namely FlatVel-A, FlatVel-B, CurveVel-A, CurveVel-B, FlatFault-A, FlatFault-B, CurveFault-A and CurveFault-B. These data reflect the variation characteristics of different geological structures and velocities. This paper extracts 10,000 velocity model samples from the eight geological structure categories of the dataset, for a total of 80,000 samples, as training data for the diffusion model. During the training process, all velocity models are clipped to a $64 \times 64$ two-dimensional grid. The maximum number of time steps in the forward diffusion process is set to 500. The Adam optimizer is used during model training, and the learning rate is set to $5 \times 10^{-4}$. The training lasts for 100 epochs.

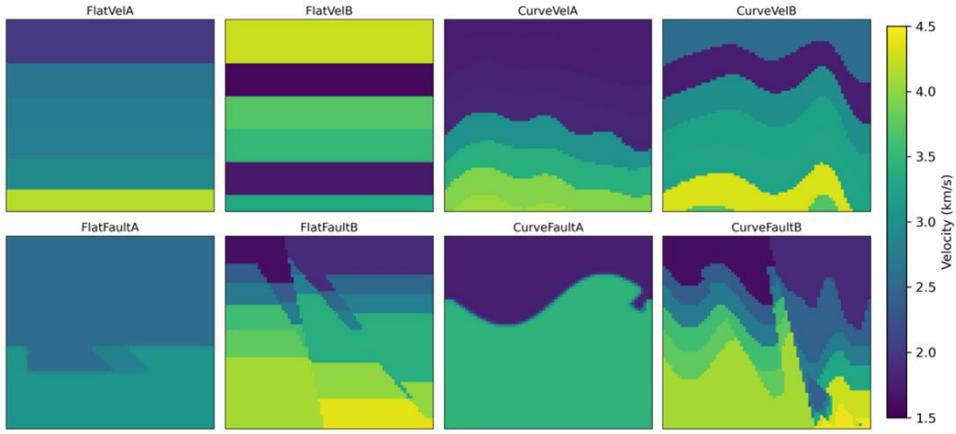

Fig. 5 Eight categories of speed model samples

## 2.2 OpenFWI Model Experiment

This section uses the velocity model shown in Figure 6 to test the performance of the proposed method. The velocity model (Figure 6(a)) follows the same velocity distribution as the training set, but is not included in the training set of the diffusion model. The initial model (Figure 6(b)) is obtained by smoothing the real velocity model. Seismic data is obtained by numerically solving the two-dimensional acoustic equation. 32 seismic sources and 64 receivers are evenly distributed on the ground. The horizontal and vertical grid spacing is set to 10 m. The seismic sources use Ricker wavelets with a dominant frequency of 15 Hz. The time sampling interval is 1 ms and the total recording time is 1.5 s. Figure 7 shows the seismic recording results of a shot gathering. Figure 7(a) is the original synthetic seismic data. In order to simulate the environmental interference in actual seismic acquisition, Gaussian white noise was added to the original data. The generated noisy synthetic seismic data is shown in Figure 7(b). After completing the pre-training of the diffusion model, comparative experiments were conducted with the unconditional diffusion regularization full waveform inversion method[19] and the traditional full waveform inversion method[22] under the conditions of no noise and noisy seismic data.

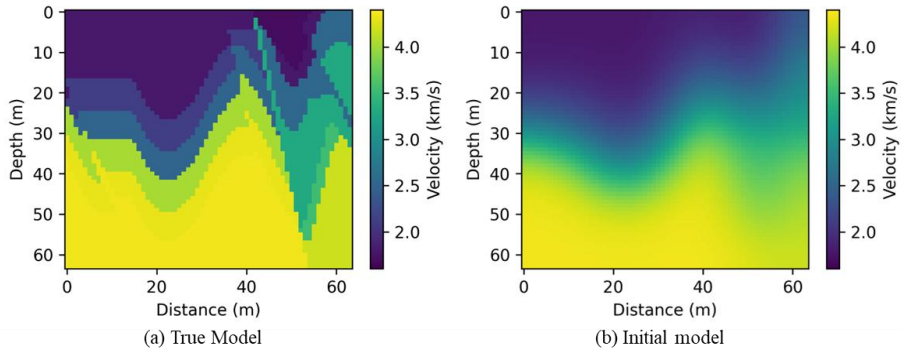

(a) True Model    (b) Initial model

Fig. 6 Velocity model and initial model

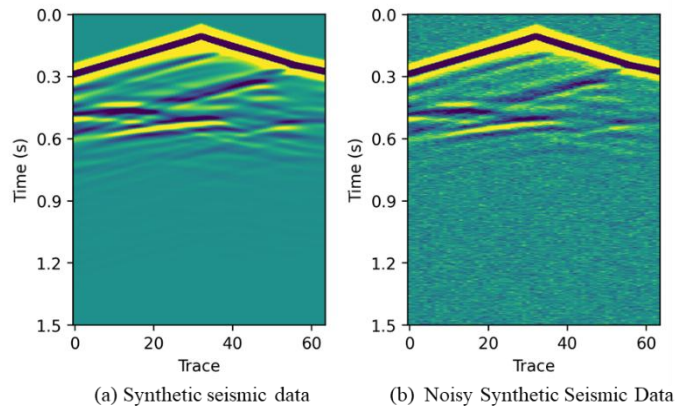

(a) Synthetic seismic data    (b) Noisy Synthetic Seismic Data

Fig. 7 Synthetic seismic data

The inversion results under noise-free seismic data conditions are shown in Figure 8. Traditional FWI results (Figure 8(a)) can reconstruct the macroscopic distribution characteristics of the subsurface velocity model relatively well, but some numerical oscillations can be observed in areas with drastic velocity changes in the middle and deep layers. Local interface transitions are slightly blurred, and there is a slight unevenness in the deep region. Unconditional diffusion-regularized FWI results (Figure 8(b)) are superior to traditional FWI results in overall structure recovery. The transition between the shallow and deep layers is smoother, the morphology of the undulating interface in the middle layer is more completely restored, interface continuity is enhanced, and the outline of the right-side tilted structure is clearer. Conditional diffusion-regularized FWI results (Figure 8(c)) perform better in terms of structural stability and detail characterization. The high-velocity area is more evenly distributed, the interface transition is more natural, and the clarity of the boundary of the undulating interface in the middle layer and the right-side tilted structure area is further improved, with stronger structural continuity.

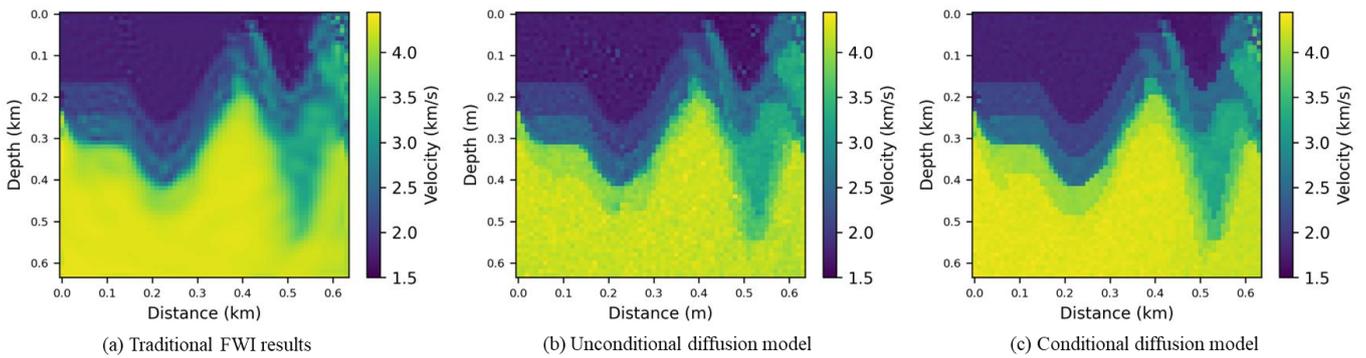

(a) Traditional FWI results  (b) Unconditional diffusion model  (c) Conditional diffusion model

Fig. 8 Inversion results of noiseless seismic data

The inversion results under noisy seismic data conditions are shown in Figure 9. Traditional FWI results (Figure 9(a)) show obvious local anomalous patches under noise, especially in areas with drastic velocity changes. The interface morphology becomes discontinuous due to noise disturbance, and artifacts and numerical oscillations appear in local areas. This indicates that traditional FWI is sensitive to noise; the random components in the data residuals directly affect the gradient direction, leading to unstable model updates and reduced reliability of the inversion results. Unconditionally diffused regularized FWI results (Figure 9(b)) show some noise resistance compared to traditional FWI. The undulating interface in the middle remains largely continuous, and the distribution of the deep high-velocity region is relatively stable. However, some speckled texture can still be observed in local areas, and interface details still fluctuate due to noise. Conditionally diffused regularized FWI results (Figure 9(c)) show the most stable noise resistance. The distribution of the deep high-velocity region is more uniform, random textures are significantly reduced, the middle interface position is clearer, and structural continuity is better. Overall, under noisy data conditions, the diffusion model regularization method significantly improves the robustness of the inversion, while the conditional diffusion model further enhances noise resistance and structure preservation by introducing physical property coupling constraints.

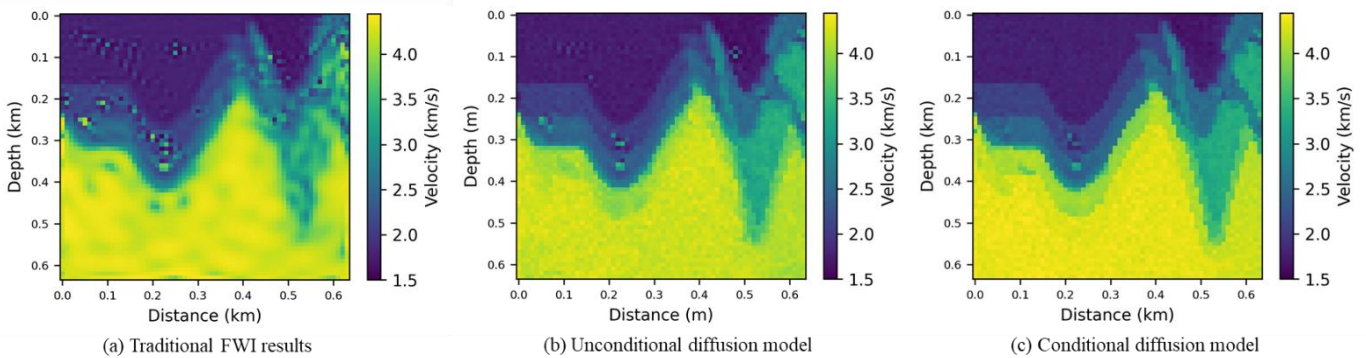

(a) Traditional FWI results  (b) Unconditional diffusion model  (c) Conditional diffusion model

Fig. 9 Inversion results of noisy seismic data

Besides the noise recorded in the seismic data, insufficient spatial sampling of the seismic source also significantly affects the full waveform inversion results. To simulate the undersampling scenario, this experiment uses only a limited number of seismic sources for forward modeling. Figure 10(a) shows the true velocity model, using only shot point data located at 140 m, 420 m, and 600 m for forward modeling to construct observation data under limited source acquisition conditions. The traditional FWI results (Figure 10(b)) show a certain degree of structural degradation under limited source conditions. The interface transition becomes blurred on both sides of the central undulating interface and near the right-side tilted structure, with slight local oscillations. The unconditional diffusion-regularized FWI results (Figure 10(c)) show a more complete undulating interface morphology than the traditional FWI results, and the outline of the right-side tilted structure is also clearer. However, in areas with severe interface undulations, the local structural boundaries are still slightly blunted. The conditional diffusion-regularized FWI results (Figure 10(d)) further improve structural recovery. The morphology of the right-side tilted structure is closer to the true model, and the local oscillation phenomenon is significantly reduced.

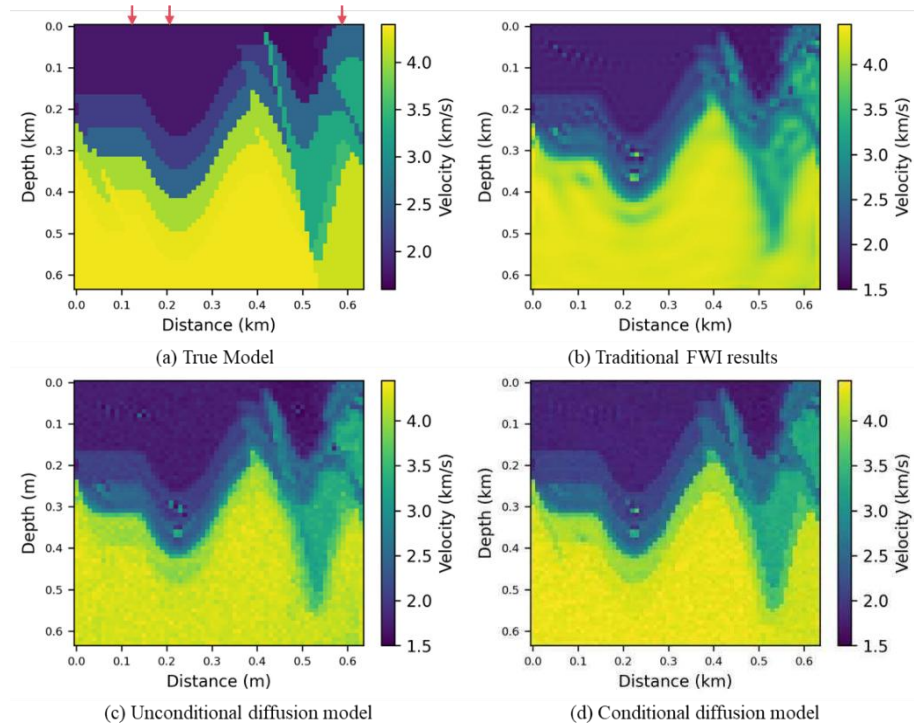

Fig. 10 Inversion result of finite shot set

To further examine its robustness and potential field application capabilities under complex real-world conditions, this experiment compares it under a more challenging initial model and low-frequency missing conditions. To simulate the low-frequency missing scenario, a high-pass filter was applied to the observation records to remove frequency components below 3 Hz. Figure 11(a) shows the initial model after further smoothing. The traditional FWI results (Figure 11(b)) exhibit significant structural distortion due to insufficient low-frequency information. The position of the central undulating interface shifts, strong oscillations and artifacts exist in local areas, and the morphology of the tilted structure on the right is not fully recovered. The unconditional diffusion-regularized FWI results (Figure 11(c)) still show local oscillations and detail deviations in some interface areas, and the stability of the inversion results is still affected by the low-frequency missing. The conditional diffusion-regularized FWI results (Figure 11(d)) show significantly reduced local oscillations and significantly enhanced structural continuity.

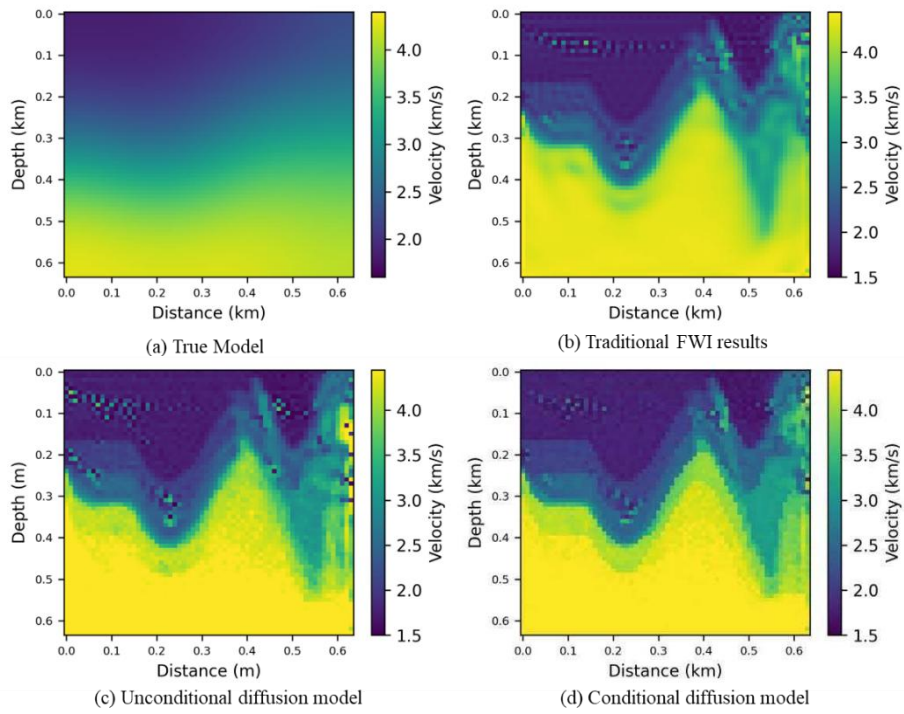

Fig. 11 Inversion results under the condition of low frequency missing

Table 5-1 presents a quantitative comparison of the three inversion methods on the MAE, MSE, and SSIM indices under four complex conditions: no noise, no noise, few shot collections, and low-frequency missing data. The table shows that the diffusion-regularized FWI method outperforms the traditional FWI in MAE, MSE, and SSIM. Compared to the unconditional diffusion model, the conditional diffusion model further improves on all indices, indicating that the introduction of density condition information enhances the diffusion model's ability to represent subsurface structures and improves the stability and robustness of the inversion results.

Table 5-1 Evaluation index of inversion results under different complex conditions

|  | Method | MAE | MSE | SSIM |
|---|---|---|---|---|
| Noise-free seismic data | Conditional diffusion | 0.0747 | 0.0202 | 0.8364 |
|  | Unconditional diffusion | 0.0955 | 0.0262 | 0.8155 |
|  | Traditional FWI | 0.1026 | 0.2633 | 0.7688 |
| Noisy seismic data | Conditional diffusion | 0.0795 | 0.0225 | 0.8083 |
|  | Unconditional diffusion | 0.1125 | 0.0292 | 0.7567 |
|  | Traditional FWI | 0.1391 | 0.5537 | 0.6281 |
| Lesser Artillery Collection | Conditional diffusion | 0.0756 | 0.0206 | 0.8201 |
|  | Unconditional diffusion | 0.1168 | 0.0294 | 0.8102 |
|  | Traditional FWI | 0.0957 | 0.2919 | 0.7455 |
| Low frequency missing | Conditional diffusion | 0.1015 | 0.0549 | 0.8180 |
|  | Unconditional diffusion | 0.1503 | 0.0716 | 0.7720 |
|  | Traditional FWI | 0.1728 | 0.2987 | 0.6911 |

2.3 Marmousi Model Experiment

To further verify the applicability and stability of the proposed method under complex geological conditions, this experiment uses the classic Marmousi velocity model for numerical experiments. The Marmousi model has complex layered structures, strong lateral velocity variations, and multiple faults and folds. The Marmousi velocity model is shown in Figure 12, and the overall velocity range of the model is approximately 1.5 km/s–5.5 km/s.

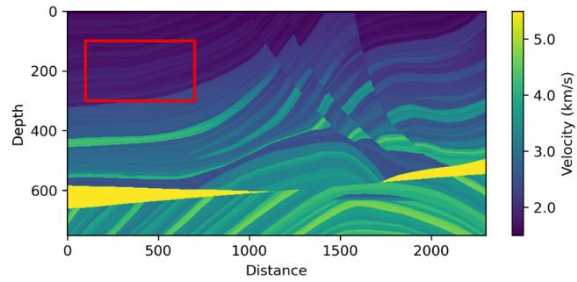

Fig. 12 Marmousi model

To improve the efficiency of the inversion calculation, this experiment extracted the region shown in the red rectangle in the figure from the original Marmousi velocity model as the research object. The selected sub-region has a grid size of 600×250 and a grid spacing of 4 m. Figure 13 shows the experimental model used for inversion, where Figure 13(a) is the selected velocity model and Figure 13(b) is the initial model obtained after spatial smoothing the real model.

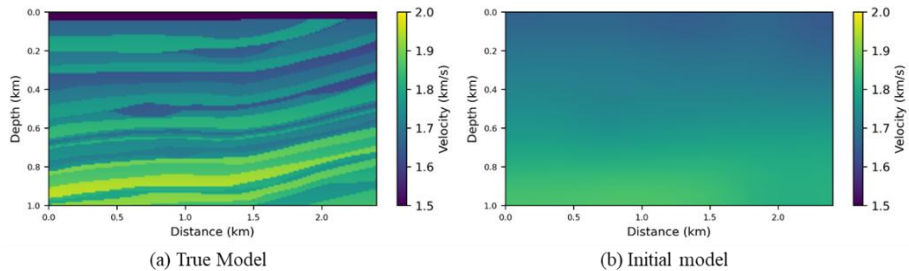

Fig. 13 Real model and initial model of Marmousi local area

The inversion results are shown in Figure 14. Traditional FWI results (Figure 14(a)) exhibit strong striped artifacts in the complex structural region on the right side of the model. This is because traditional FWI is susceptible to multipath propagation and nonlinearity under complex conditions, leading to unstable gradient updates. The unconditional diffusion model results (Figure 14(b)) show better stability compared to traditional FWI, with reduced oscillations in the complex structural region on the right, but some complex structural details are still not fully recovered. The conditional diffusion model results (Figure 14(c)) further improve structural recovery. Compared to the unconditional diffusion model, the overall layered structure of the model is clearer, the continuity of the complex structural region on the right is significantly enhanced, and local oscillations are further reduced. The results indicate that by introducing density condition information, the diffusion model can better characterize the coupling relationship between different physical parameters in the subsurface medium, thus providing more effective prior constraints under complex geological conditions and improving the stability and structural consistency of the inversion results.

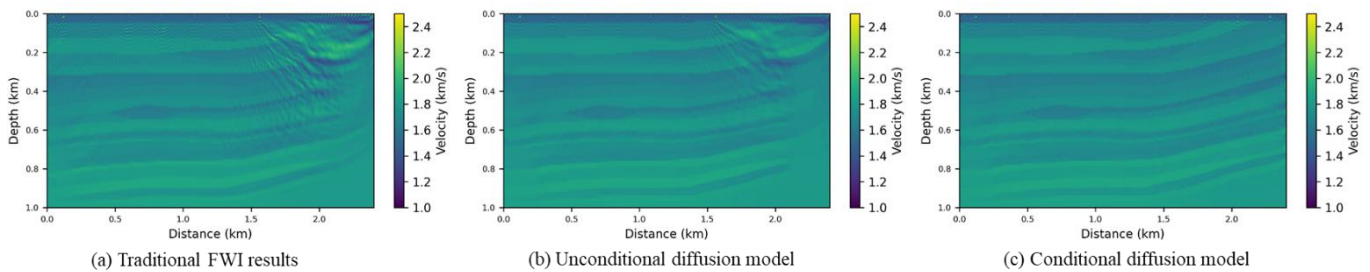

Figure 14 Local Area Inversion Results

## 3 Conclusion

This paper improves the diffusion model structure based on existing diffusion-regularized full-waveform inversion methods by introducing two-dimensional density information as a conditional constraint to construct a conditional diffusion model. This enhances the ability of prior distributions to represent the structural characteristics of subsurface media. Experimental results on the OpenFWI dataset and the Marmousi model show that, with the introduction of density conditional information, the diffusion model can better characterize the coupling relationships between different physical properties in the subsurface media, thereby further improving the accuracy and stability of the inversion results. This method provides an effective approach to improving the performance of full-waveform inversion under complex conditions and has certain application potential.。

# References

[1] Kenneth Aki, Paul G. Richards.Quantitative Seismology: Theory and Methods.W.H. Freeman, 1980.

[2] Norton S J. Iterative seismic inversion[J]. Geophysical Journal International, 1988, 94(3): 457–468.

[3] Lailly P. The seismic inverse problem as a sequence of before stack migrations[C]// Bednar J B, Robinson E, Weglein A, eds. Conference on Inverse Scattering—Theory and Application. Philadelphia: SIAM, 1983: 206–220.

[4] Tarantola A. Inversion of seismic reflection data in the acoustic approximation[J]. Geophysics, 1984, 49(8): 1259–1266.

[5] Pratt R G, Shipp R M. Seismic waveform inversion in the frequency domain; Part 2: fault delineation in sediments using crosshole data[J]. Geophysics, 1999, 64(3): 902–914.

[6] Shin C, Cha Y H. Waveform inversion in the Laplace–Fourier domain[J]. Geophysical Journal International, 2009, 177(3): 1067–1079.

[7] Xu S, Wang D, Chen F, et al. Inversion on reflected seismic wave[C]// SEG International Exposition and Annual Meeting. Tulsa: Society of Exploration Geophysicists, 2012.

[8] Leonid Rudin, Stanley Osher, Emad Fatemi.Nonlinear total variation based noise removal algorithms.Physica D, 1992, 60(1–4): 259–268.

[9] Felix J. Herrmann, Xiaolei Li.Efficient least-squares imaging with sparsity promotion and compressive sensing.Geophysical Prospecting, 2012, 60(4): 696–712.

[10] Haitao Zhu, Sergey Fomel.Full waveform inversion with dictionary learning regularization.Geophysics, 2017, 82(5): R309–R321.

[11] Tariq Alkhalifah.Full-model wavenumber inversion: An efficient solution for full waveform inversion.Geophysics, 2018, 83(6): R583–R593.

[12] Gongsheng Huang, Sergey Fomel.Robust full-waveform inversion using dictionary learning regularization.Geophysics, 2019, 84(6): R1045–R1059.

[13] Yue Li, Tariq Alkhalifah, Zhen Zhang.Deep-learning assisted regularized elastic full waveform inversion using velocity distribution information from wells.Geophysical Journal International, 2021, 226(2): 1322–1335.

[14] Sheng Li et al.Deep-learning inversion of seismic data.IEEE Transactions on Geoscience and Remote Sensing, 2020, 58(3): 2135–2149.

[15] Mosser L, Dubrule O, Blunt M J. Stochastic seismic waveform inversion using generative adversarial networks as a geological prior[J]. Mathematical Geosciences, 2020, 52(1): 53–79.

[16] Zhang W, Gao J H. Deep-learning full-waveform inversion using seismic migration images[J]. IEEE Transactions on Geoscience and Remote Sensing, 2021, 60: 1–18.

[17] Tariq Alkhalifah, Hao Wang, Oleh Ovcharenko.MLReal: Bridging the gap between training on synthetic data and real data applications in machine learning.Artificial Intelligence in Geosciences, 2022, 3: 101–114.

[18] Yang F S, Ma J W. FWIGAN: Full-waveform inversion via a physics-informed generative adversarial network[J]. Journal of Geophysical Research: Solid Earth, 2023, 128(4): e2022JB025493.

[19] Wang F, Huang X, Alkhalifah T A. A Prior Regularized Full Waveform Inversion Using Generative Diffusion Models.[J] IEEE Transactions on Geoscience and Remote Sensing, 2023, vol. 61, pp. 1-11.

[20] Jonathan H, Ajay Jain, Pieter Abbeel. Denoising Diffusion Probabilistic Models[J]. 2020. DOI:10.48550/arXiv.2006.11239.

[21] Deng C, Liu Y, Zhang J, et al. OpenFWI: Large-scale multi-structural benchmark datasets for seismic full waveform inversion[EB/OL]. arXiv:2111.02926, 2021.

[22] A. Richardson, "Deepwave," Zenodo, 2022, doi: 10.5281/zen-odo.7278382.

Author Biographies:
Liu Caiyun, female, born in 1975, PhD, Associate Professor, major research interests: geophysical inversion, artificial intelligence; Email: 100513@yangtzeu.edu.cn.
Corresponding Authors:
Yu Qingfeng, female, born in 2001, Master's student, Information and Communication Engineering; Email: 2023710658@yangtzeu.edu.cn.
Xiong Jie, male, born in 1975, PhD, Professor, major research interests: artificial intelligence, geophysical inversion theory; Email: xiongjie@yangtzeu.edu.cn